\documentclass[10pt,twocolumn,letterpaper]{article}

\usepackage{cvpr}              

\usepackage[dvipsnames]{xcolor}

\definecolor{cvprblue}{rgb}{0.21,0.49,0.74}
\usepackage[pagebackref,breaklinks,colorlinks,citecolor=cvprblue]{hyperref}

\def\confName{CVPR}
\def\confYear{2024}

\title{2nd Place Solution for CVPR2024 E2E Challenge: End-to-End Autonomous Driving Using Vision Language Model}

\author{Zilong Guo, Yi Luo, Long Sha, Dongxu Wang, Panqu Wang, Chenyang Xu, Yi Yang \\
ZERON  \\
Shanghai, China \\
{\tt\small panqu.wang@zeron.ai}
}

\begin{document}
\maketitle
\begin{abstract}
End-to-end autonomous driving has drawn enormous attention recently. Many works focus on the use of modular deep neural networks to construct the end-to-end architecture. However, whether using powerful large language models (LLM), especially multi-modality Vision Language Models (VLM) could benefit the end-to-end driving tasks remains a question. In our work, we demonstrate that the combination of end-to-end architectural design and knowledgeable VLMs yields impressive performance on driving tasks. It should be noted that our method only uses a \textbf{single} camera and is the best camera-only solution on the leaderboard \footnote{\url{https://opendrivelab.com/challenge2024/\#end_to_end_driving_at_scale}}, demonstrating the effectiveness of the vision-based driving approach and the potential for end-to-end driving tasks. \footnote{For questions and inquiries, please direct to \href{mailto:panqu.wang@ezeron.ai}{panqu.wang@zeron.ai}}
\end{abstract}    
\section{Introduction \& Related Work}
\label{sec:intro}

Autonomous driving technology has been evolving rapidly. Mainstream architecture involves dividing the whole pipeline into several different functions, such as perception, localization \& mapping, prediction, planning, and control. This process has been considered to be a decent trade-off between performance, safety, and explainability. However, while the driver assistance system (L2 to L3) starts to occupy a large market share, high-level (L4+) autonomous driving still lags behind mass production. There are several reasons behind it:

\textbf{Complex architecture}: The current mainstream solution has twenty or even more modules. Due to computing power limitations, the performance ceiling of a single module is not high; there are too many internal interfaces in the system, and transmission and optimization are difficult; local and overall optimization goals sometimes conflict, so the performance improvements is hard to observe.

\textbf{High cost}: R\&D/maintenance/manpower costs soar as the number of modules increases. Repeated reinvention, ineffective organization and integration problems often occur. Overall project coordination, iteration and management are usually difficult.

\textbf{Poor generalization}: The under-performed system usually incurs so-called  long-tail problems. If not properly handled, or as time elapses, the overwhelming number of handcrafted solutions could result in poor maintainability and scalability.

\textbf{Difficult to Commercialize}: Current high-level autonomous driving products could only be used in limited scenarios such as small number of cities/demonstration areas/highways. The algorithm is generally deeply associated with software and hardware platforms, and it is also difficult to be made compatible version with more vehicle models/platforms/scenarios.

Compared with the traditional architecture, end-to-end autonomous driving generally has the following advantages:

\textbf{Simplified Framework}: The entire system is encapsulated within a single module, leveraging an end-to-end deep neural network for a high integration level. We could also share the same framework for both online and offline environments, enhancing development efficiency and maintenance friendliness.

\textbf{Cost Efficiency}: End-to-end solution achieves a large amount of cost reduction and efficiency enhancement. It lowers costs associated with algorithm development, testing, deployment, module iteration, data feedback loops, and project management. It also accelerates the product deployment and customer service effectiveness.

\textbf{Strong Generalization}: End-to-end autonomous driving combines data-driven and knowledge-driven approaches for a comprehensive understanding of the world. It seamlessly tackles long-tail problems, yielding more robust model outputs and facilitating easier customization of requirements, thereby enhancing the efficiency of product iterations.

\textbf{Production-Friendly}: It achieves excellent performance on many benchmarks. A single adaptable architecture swiftly accommodates diverse scenarios, vehicle models, and platforms, minimizing marginal costs and enabling advanced autonomous driving technology to become widely accessible to the society.

There has been many innovative and influential works on end-to-end autonomous driving. A main line of literature \citep{hu2022st, hu2023planning, chen2024vadv2, zhang2024graphad} employs modularized deep neural networks to learn effective intermediate features (sometimes including detection/tracking/mapping/prediction/scene understanding sub-tasks) to generate the planning trajectory and/or control signal at the end. Meanwhile, considerable number of works \citep{mao2023gpt, tian2024drivevlm,wang2023drivemlm,xu2023drivegpt4,mao2023language,shao2023lmdrive,ding2024holistic,wang2024omnidrive} introduce large language models (LLMs) into the end-to-end modeling framework and showcases excellent performance on driving tasks, even including Q\&A problems. Recently, generative modeling on autonomous driving tasks has also drawn great attention and yields impressive effects \citep{hu2023gaia,li2023drivingdiffusion,lu2023wovogen,wang2023driving,ding2024holistic,zhao2024drivedreamer,yang2024generalized,gao2024vista}. In our work, we absorb the aforementioned works and designing an end-to-end multi-modality language model-based solution for autonomous driving. Remarkably, our whole system only uses the input from a single camera and still achieves top performance on the leaderboard.

\section{Method}

\begin{figure*}[h]
  \begin{center}
    \includegraphics[width=1.0\textwidth]{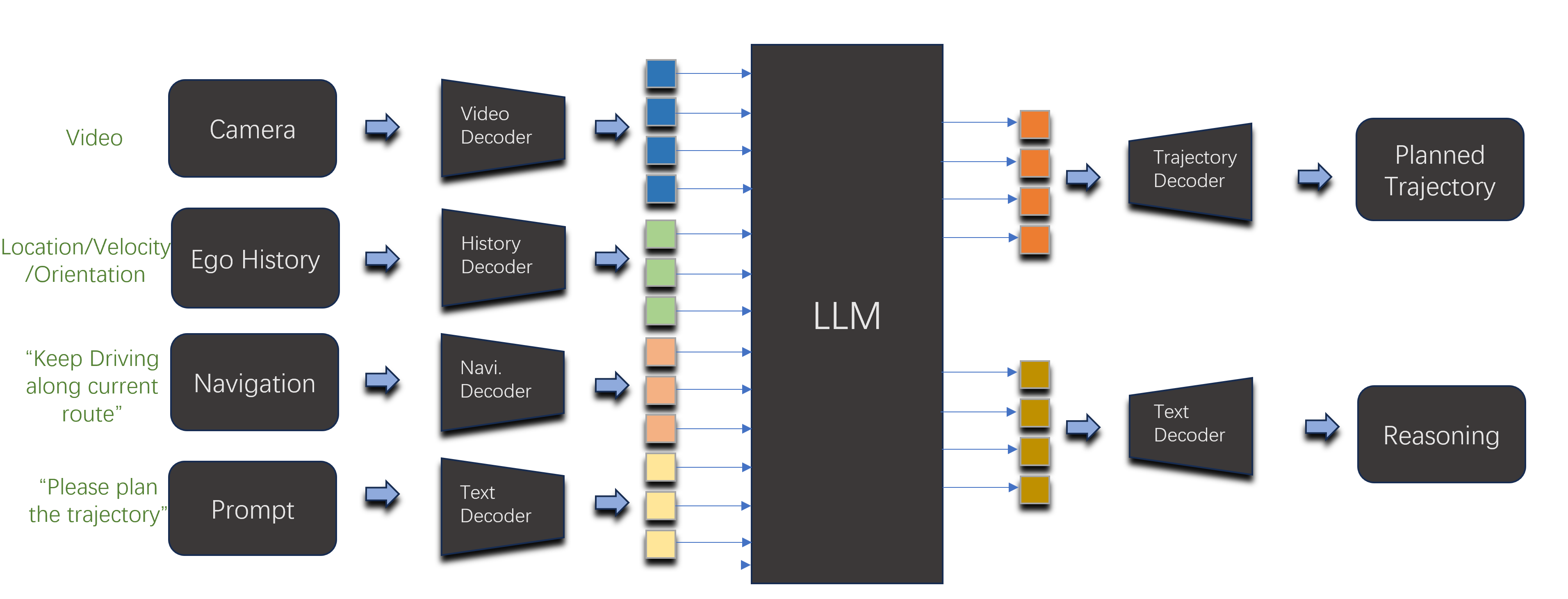}
  \end{center}
  \caption{\textbf{The architecture of our network.} Our network receives input from camera, ego history, navigation signal, and text prompt. Through various encoder and the LLM module, our network generates trajectories and texts simultaneously.}
  \label{fig:architecture}
\end{figure*}

Figure \ref{fig:architecture} shows the architecture of our network. As an end-to-end model, our network directly receives information from raw sensor (camera in our case), indicating "what we have seen". It also obtains the ego history information to understand "what have we done". The navigation signal explains "where to go next". The flexible text prompt conveys "what to do in this task". 

\textbf{Encoder} All inputs are followed by the corresponding encoder, either vision transformer \citep{dosovitskiy2020image} for the camera image, standard MLP for the ego history (represented by location, velocity, and acceleration vector), or text tokenizer for the navigation command (represented by pure text command such as "go straight for 50 meters", or "turn left now") and text prompt (such as "please generate a safe trajectory for next 5 seconds"). 

\textbf{LLM} For the LLMs, any open-source model 
 such as \citep{touvron2023llama, cai2024internlm2, vicuna} works, depending on the computational resources. We use the Intern-LM 4B \cite{cai2024internlm2} model throughout our experiments. 
 
 \textbf{Decoder} LLm generates the output tokens autoregressively. For the trajectory decoder, we simply use an MLP to decode the trajectory vector which is represented by location, velocity, and acceleration based on the last token. The text output is generated after the trajectory so it is consistent with the trajectory representation when the ground truth is available. 
 
 \textbf{Training} We train the network for multiple epochs on the provided nuPlan \citep{caesar2021nuplan} dataset, using LoRA \citep{hu2022lora}. The trajectory output is trained using L2 loss, and the text output is trained using next token prediction loss. The text output is also trained using driving language dataset such as \citep{sima2023drivelm}, which provides valuable training data for the network to understand the environment. 
 
 Although the design is simple, it resembles how the human drives vehicle in reality: By using eyes, receiving navigation and goal information, human can plan the trajectory in next few seconds and infer why should we execute the trajectory simultaneously. We believe modeling driving like human is a promising research direction, and may yield fruitful results in the future.

\section{Result}

 We validate our model on the CVPR 2024 End-to-End Driving at Scale Challenge track. We achieve the final score of 0.8747, and is the best camera-only solution across the leaderboard. Since we just use one front-facing camera with single-frame input, this result suggest huge potential for the proposed framework. 
 
\begin{figure*}[h]
  \begin{center}
    \includegraphics[width=1.0\textwidth]{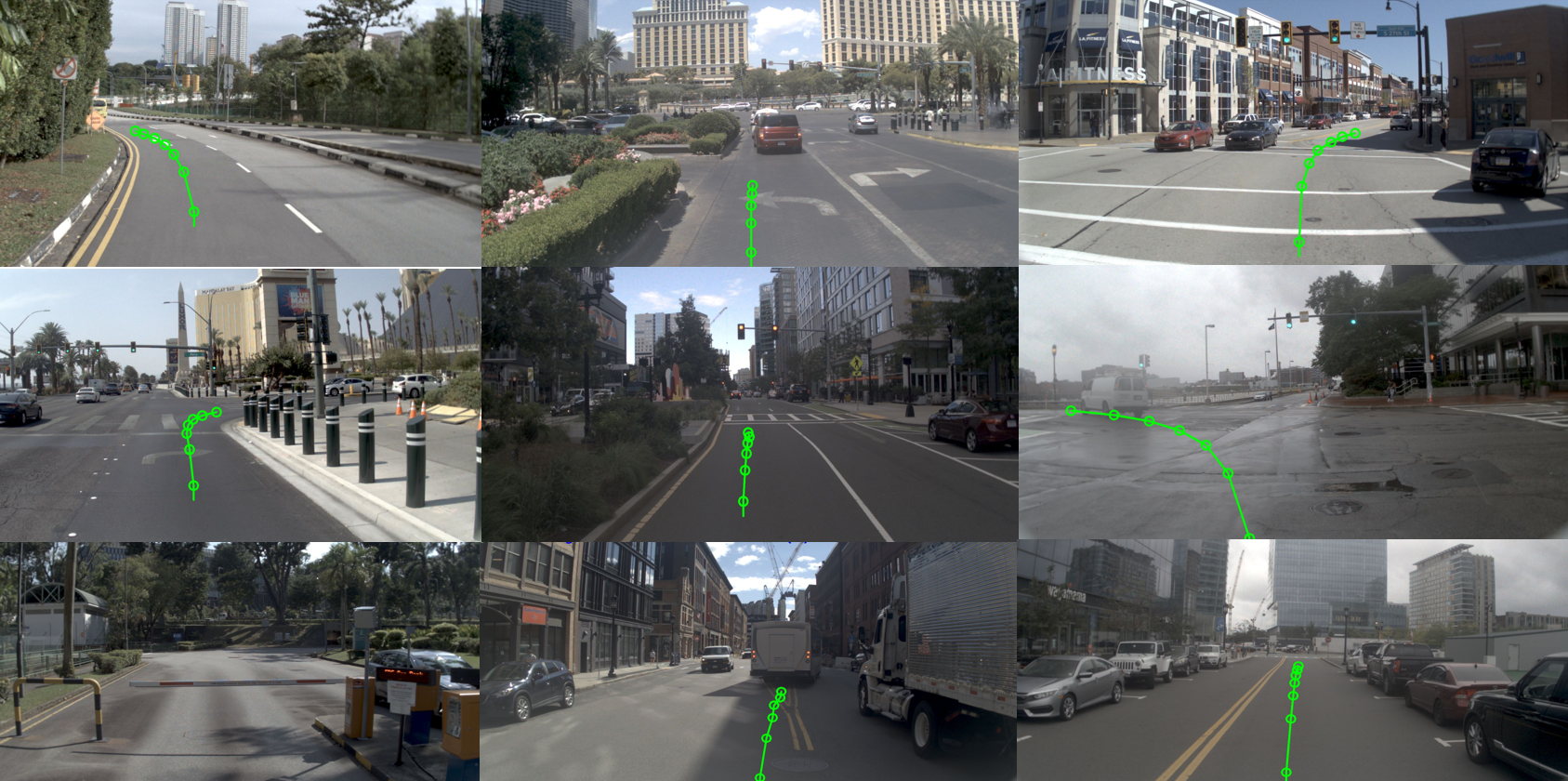}
  \end{center}
  \caption{\textbf{Visualization of the results of our method.}}
  \label{fig:vis}
\end{figure*}

 Some visualizations could be seen in Figure \ref{fig:vis}. We can see that the model handles various driving scenarios very well, such as going straight, left/right turn, waiting traffic light, stop sign, and even corner cases like toll booth. Specifically, we test the model to see if it just extrapolate the history trajectory by looking at some stop and go conditions such as waiting at traffic light, stop sign, going through toll-booth, swerving through parked vehicles, and waiting for pedestrians. The result also suggests the model learns the human-like driving behavior under such scenarios. The good performance and potential strong generalization power suggests that the model does have high performance potential and leads towards a promising approach for autonomous driving production. 
 
 \textbf{Limitations} Due to time and resource limitation, we did not conduct comprehensive ablation studies. The single-frame single-camera approach is quite primitive in the real-world setting, and we need to integrate the multi-sensor multi-frame setting into the framework, which may require careful design for the visual encoder, 3D spatial representation, and LLM training paradigm. In addition, the open-loop setting may not reveal the real-world or close-loop performance, and the corresponding testing environment should also be developed. However, we demonstrate that the proposed end-to-end architecture naturally fits the autonomous driving tasks and have huge potential of achieving even better performance, given more data, computational resource, and sensory input. We plan to explore these directions in the future.

{
    \small
    \bibliographystyle{ieeenat_fullname}
    \bibliography{main}
}


\end{document}